\documentclass[cameraready]{Interspeech}
\usepackage{newtxtext}
\usepackage{amsmath,amsfonts,bm}

\def\eqref#1{equation~\ref{#1}}
\def\1{\bm{1}}

\DeclareMathAlphabet{\mathsfit}{\encodingdefault}{\sfdefault}{m}{sl}
\SetMathAlphabet{\mathsfit}{bold}{\encodingdefault}{\sfdefault}{bx}{n}

\def\code#1{\texttt{#1}}
\usepackage{xspace}

\usepackage{array, boldline, makecell, booktabs}
\usepackage[svgnames, table]{xcolor}

\makeatletter
\NewDocumentCommand{\supptitle}{s}{
\twocolumn[{
  \begin{center}
    \vspace*{-0.3cm}
    \rule{\textwidth}{0.05cm}\\[0.1cm]
    \textbf{- Appendix -}\\[0.2cm]
    {\Large \textbf{\mytitle}}\\[0.1cm]
    \rule{\textwidth}{0.05cm}\\[0.3cm]
  \end{center}
}]
}
\makeatother

\newcommand{\eg}{\emph{e.g.,~}}
\newcommand{\ie}{\emph{i.e.,~}}

\newcommand{\mc}[1]{\mathcal{#1}}

\definecolor{LightGray}{rgb}{0.88,0.88,0.88}
\definecolor{LightCyan}{rgb}{0.88,1,1}
\definecolor{Blue}{rgb}{0, 0.5, 1}
\definecolor{Green}{rgb}{0.0, 0.8, 0.0 }
\definecolor{Red}{rgb}{0.95, 0.55, 0.6}
\definecolor{Skyblue}{rgb}{0.6, 0.6, 0.95 }
\definecolor{Beige}{rgb}{0.96, 0.96, 0.86}

\newcommand{\alg}{\textbf{\textsc{STeReO}}\xspace}
\newcommand{\mytitle}{A Reranker for Orchestrating Heterogeneous Speech and Text Retrievers}

\usepackage[normalem]{ulem}
\useunder{\uline}{\ul}{}

\usepackage{tcolorbox}
\tcbuselibrary{skins, breakable}

\usepackage[hang,flushmargin]{footmisc}

\newcommand{\myparagraph}[1]{\vspace{0.07cm}\noindent\textbf{#1}~}

\usepackage{booktabs}
\usepackage{tabularx}
\usepackage[table]{xcolor}
\usepackage{multirow}
\usepackage{array}
\usepackage{amsmath, amssymb, amsfonts}
\usepackage{graphicx}
\usepackage{placeins}
\usepackage{cite}
\usepackage{float}
\usepackage[ruled,vlined,linesnumbered]{algorithm2e}

\newcommand{\sys}[1]{\textsc{#1}}


\title{\mytitle}

\author[affiliation={1}]{Inho}{Kim}
\author[affiliation={1}, correspondingauthor]{Sumyeong}{Ahn}

\address{
    $^1$ Korea Institute of Energy Technology, South Korea
}

\email{inho20@kentech.ac.kr, sumyeongahn@kentech.ac.kr}

\begin{document}
\maketitle

\begin{abstract} 
Retrieval-Augmented Generation (RAG) systems have attracted significant interest for their ability to mitigate hallucinations in Large Language Models (LLMs). Although knowledge databases for RAG are increasingly diversifying to include various modalities such as speech and text, research on handling such multi-modal database scenarios remains limited. In this paper, we propose \alg (\textbf{\underline{S}}peech and \textbf{\underline{Te}}xt \textbf{\underline{R}}eranking \textbf{\underline{O}}rchestrator), a reranker based on speech and text retrievers that aggregates disparate modality databases. To address the lack of specialized training data, we first curate a dataset comprising queries, mixed-modality evidence, and their corresponding relevance ranks. We then train the reranker and evaluate its effectiveness in both single-modality and mixed-modality scenarios. Our results demonstrate that the proposed algorithm excels at selecting the most relevant evidence, thereby significantly improving downstream question-answering performance. 
\end{abstract}

\section{Introduction}

\begin{figure}[t]
    \centering
    \includegraphics[width=1.15\columnwidth]{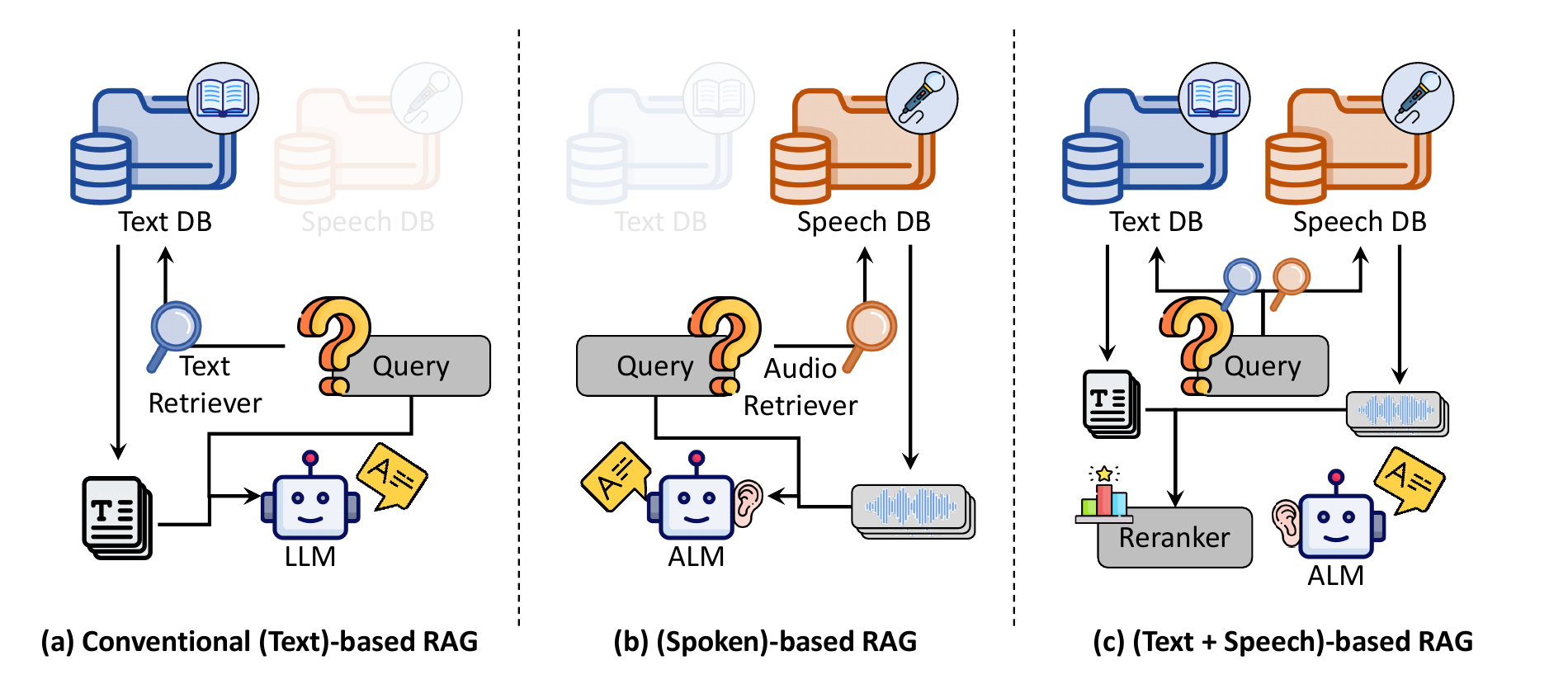}
    \vspace{-10pt} 
    \caption{Comparison of three RAG systems: (a) Text-based, (b) Speech-based, and (c) Text+Speech-based RAG.}
    \label{fig:intro}
    \vspace{-20pt}
\end{figure}

Retrieval-Augmented Generation (RAG)~\cite{lewis2020retrieval} enhances Large Language Models (LLMs)~\cite{achiam2023gpt, team2023gemini, chu2024qwen2} by incorporating external knowledge, thereby mitigating hallucination~\cite{huang2025survey} and grounding responses in external evidence. While most existing RAG pipelines assume text-only knowledge bases (\autoref{fig:intro}(a)), there is a growing need to integrate  unstructured spoken content such as lectures and meeting recordings. However, this integration is non-trivial. The most straightforward approach is leveraging Automatic Speech Recognition (ASR), which converts audio to text before applying standard text retrieval. However, ASR introduces transcription errors that propagate through the pipeline, adds latency, and discards non-verbal information present in the original audio. Consequently, extending RAG to natively support spoken modalities is a significant, yet largely unexplored, challenge.

Several studies address ASR-related limitations through ASR-free speech retrieval methods. For instance, \code{VoxRAG}~\cite{rackauckas2025voxrag} employs direct speech-to-speech matching without ASR, while SpeechRAG~\cite{min2024speech} aligns textual queries with speech embeddings. However, their retrieval spaces are still restricted to single-modality, speech-only corpora (\autoref{fig:intro}(b)). 
In contrast, \code{WavRAG}~\cite{chen2025wavrag} facilitates heterogeneous retrieval by projecting independent audio and text databases into a shared embedding space.
Despite this advancement, such joint embedding approaches for multimodal bases suffer from the well-known \emph{modality gap}~\cite{liang2022mind,feng2025enhancing}. Even in audio-text models like CLAP~\cite{elizalde2023clap}, systematic score imbalance~\cite{saijo2025leveraging} often leads one modality to dominate retrieval results, regardless of its actual relevance.

An alternative to avoid this gap is to employ modality-specific retrievers independently and merge their candidates via late fusion (\autoref{fig:intro}(c)). In this paradigm, a robust cross-modal reranker is essential for accurately evaluating and prioritizing candidates retrieved from diverse modalities. While recent listwise reranking methods, such as those based on permutation-invariant cross-encoders~\cite{schlatt2025set} or Fusion-in-Decoder architectures~\cite{yoon2024listt5}, have shown promising results on text-only pools, they remain inherently unimodal. Consequently, they fail to facilitate the complex cross-modal comparison necessitated by heterogeneous candidate pools. A primary hurdle in developing such a cross-modal reranker is the absence of training data: existing retrieval datasets lack explicit cross-modal relevance judgments required to directly compare audio and text candidates.

To overcome this challenge, we first construct a novel dataset that explicitly captures the cross-modal relevance rankings among candidates  generated by heterogeneous retrievers. Leveraging this dataset, we propose \alg (\textbf{\underline{S}}peech and \textbf{\underline{Te}}xt \textbf{\underline{Re}}ranking \textbf{\underline{O}}rchestrator), a cross-modal reranker designed to systematically align and prioritize heterogeneous candidates --drawn from independent, modality-specific databases-- into a single, unified ranked list for a given text query. In summary, our main contributions are as follows:

\begin{itemize}
    \item First, we construct a novel cross-modal dataset that provides explicit relevance rankings for candidates retrieved from heterogeneous modalities. To achieve this efficiently, we propose a novel score fusion method that merges candidate sets from disjoint retrievers. This approach effectively filters the vast evidence space, allowing us to accurately extract relevance orders from a highly targeted subset of candidates.

    \item Second, leveraging the constructed dataset, we train a novel cross-modal reranker, \alg. It is fine-tuned using Low-Rank Adaptation (LoRA) and operates via a unified token-based scoring mechanism. Furthermore, the rich annotations within our dataset enable the proposed algorithm to be optimized across various learning objectives, including pointwise, pairwise, and listwise reranking formulations.

    \item Finally, we comprehensively evaluate \alg on a cross-modal benchmark constructed from the Spoken SQuAD and MS MARCO datasets. Extensive experiments across various backbone architectures demonstrated that our proposed method outperforms existing single-modality baselines. 

\end{itemize}

\section{The proposed method \alg}
\label{sec:method}

In this section, we present the proposed method, detailing the construction of a dataset equipped with ranking labels across heterogeneous modalities.

\begin{figure}[t]
    \centering
    \includegraphics[width=1.0\columnwidth]{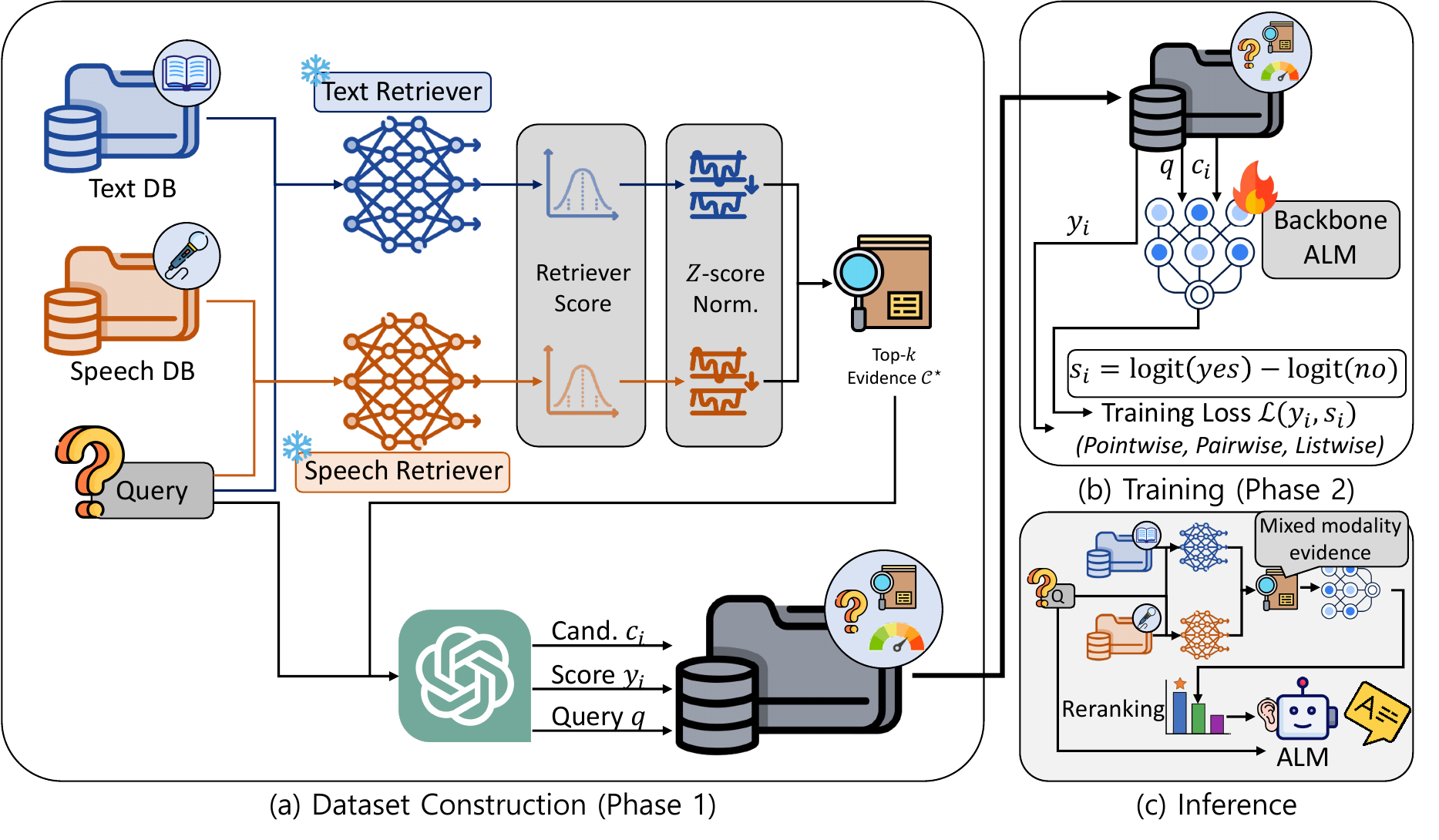}
    \vspace{-12pt} 
    \caption{Overview of the proposed framework \alg. (a) construction dataset for mixed modality reranker, (b) training reranker, and (c) inference procedure based on the proposed reranker model.}
    \vspace{-7pt} 
    \label{fig:overview}
\end{figure}

\myparagraph{Framework.} Prior to detailing the proposed method, we establish the core framework of this study. Let $q$ denote a text query and $\mc{D}^{m}$ represent heterogeneous databases, where $m \in M = \{\text{text}, \text{speech}\}$. For each modality $m$, a modality-specific retriever $R^m$ extracts a candidate set $\mc{C}^m = \{c_1^m, \ldots, c_k^m\}$, where $k$ is the number of candidates from the retriever. Subsequently, a reranker $\mc{R}$ processes the union of these sets to yield the mixed evidence $\mc{C}^{\star} = \mc{R}(\bigcup_{m\in M} \mc{C}^m)$. Finally, the Audio Language Model (ALM) generates the answer $a$ to the query as $a = \text{ALM}(\mc{C}^{\star}, q)$. The inference procedure is described in~\autoref{fig:overview}~(c). Here, our primary focus is on training the reranker module $\mc{R}$. To achieve this, we first construct the necessary training dataset (Phase 1), followed by the formal training of $\mc{R}$ (Phase 2).

\subsection{Phase 1: Dataset Construction}
In Phase 1, we construct a cross-modal dataset by aggregating candidates from independent, modality-specific retrievers and annotating them with unified relevance labels using a foundation ALM, \eg \code{GPT} (\autoref{fig:overview}~(a)).
%  \eg \code{GPT}. The entire procedure is described in~\autoref{fig:overview}~(a).

\myparagraph{Domain-specific Retrieving.}
We first generate a candidate set $\mc{C}^m$ for each modality and compute initial relevance scores. To ensure high-quality retrieval in each domain, we employ specialized pre-trained retrievers: \code{e5-mistral-7b-instruct}~\cite{wang2023e5mistral} for text and a \code{HuBERT}-based \code{SpeechRAG}~\cite{min2024speech} for speech.

\myparagraph{Normalization and Fusion.}
Since independent retrievers operate on disparate score scales, direct comparison is infeasible. To facilitate efficient candidate selection and minimize downstream labeling costs, we align these scores using $Z$-normalization:
\begin{equation*}
    \tilde{r}_i = \frac{r_i - \mu_m}{\sigma_m} \quad \text{for} \quad m \in M,
\end{equation*}
Here, $r_i$ and $\tilde{r}_i$ represent the raw and normalized retrieval scores for the $i^{\text{th}}$ candidate, while $\mu_m$ and $\sigma_m$ denote the mean and standard deviation of scores for modality $m$ given a query $q$. We then merge the candidates based on these normalized scores to form a unified labeling set $L$, from which the top-$k$ samples are selected.

\myparagraph{Cross-modal Labeling via Foundation ALM.}
To establish a gold-standard ranking across different modalities, we utilize a foundation ALM (\eg \code{gpt-4o-audio-preview}) as a unified evaluator. The top-$k$ candidates in $L=\{c_1, \ldots, c_k\}$ are fed into the ALM, which assigns individual relevance scores $\textbf{y}=\{y_1,\ldots,y_k\}$ by considering both audio and text contexts simultaneously. Detailed prompts and implementation specifications are provided in~\autoref{sec:setup} and the Supplementary Material.

\subsection{Phase 2: Reranker Training}
In phase 2, we fine-tune the reranker using the cross-modal candidate-relevance tuples $(q, c_i,y_i)$ generated in the previous phase. To ensure parameter efficiency, we employ Low-Rank Adaptation (LoRA)~\cite{hu2022lora}. The training process is depicted in~\autoref{fig:overview}~(b).

\myparagraph{Autoregressive Relevance Scoring.}
We adopt decoder-based ALMs, such as \sys{Ultravox}~\cite{fixie2024ultravox}, \sys{Qwen-Audio-Chat}~\cite{chu2023qwenaudio}, and \sys{Qwen2-Audio}~\cite{chu2024qwen2}, as our base architecture for reranker. Following established autoregressive reranking paradigms~\cite{sun2023chatgpt, qin2024large}, the model is trained to generate a scalar relevance score $s_i$, based on the logit difference between the tokens \code{Yes} and \code{No} at the final token position:
\begin{equation*}
    s_i = \text{logit}(\code{Yes})-\text{logit}(\code{No}).
\end{equation*}

\begin{table}[h]
\centering
\caption{Training objectives for reranker optimization. $\sigma(\cdot)$ denotes the sigmoid, and $\tau$ is a hyperparameter set to $10$.}
\label{tab:objective}
\renewcommand{\arraystretch}{1.2} 
\resizebox{\columnwidth}{!}{
\begin{tabular}{l l}
\toprule
\textbf{Method} & \multicolumn{1}{c}{\textbf{Objective Formulation}} \\
\midrule
\textbf{Pointwise} & 
$\displaystyle \mathcal{L}_{\text{point}} = -\frac{1}{|\mc{C}^{\star}|}\sum_{i} \bigl[ y_i \log \sigma(s_i) + (1{-}y_i)\log(1{-}\sigma(s_i)) \bigr]$ \\

\textbf{Pairwise} & 
$\displaystyle 
\begin{aligned}[t] 
    \mathcal{L}_{\text{pair}} = & \frac{1}{|\mc{P}|} \sum_{(i,j) \in \mathcal{P}(q)} \log\left(1+\exp\left(-\delta_{ij} \cdot (s_i - s_j)\right)\right) \\
    &\text{where} \quad \mathcal{P} = \{(i,j) : c_i, c_j \in \mc{C}^{\star}, y_i \neq y_j\}, \quad \delta_{ij} = \operatorname{sign}(y_i - y_j)
\end{aligned}$ \\

\textbf{Listwise} & 
$\displaystyle 
\begin{aligned}[t] 
    % \mathcal{L}_{\text{list}} = 1 - \frac{1}{\mathrm{IDCG}} \sum_i &\frac{2^{y_i}-1}{\log_2(1+\hat{r}_i)} \\
    % & \text{where} \quad \hat{r}_i = 1 + \sum_{j \neq i}\sigma\left(\tau(s_j - s_i)\right)
    \mathcal{L}_{\text{list}} = 1 - \frac{1}{\mathrm{IDCG}} \sum_i \frac{2^{y_i}-1}{\log_2(1+\hat{r}_i)}  \quad
     \text{where} \quad \hat{r}_i = 1 + \sum_{j \neq i}\sigma\left(\tau(s_j - s_i)\right)
\end{aligned}$ \\
\bottomrule
\end{tabular}%
}
\end{table}

\begin{table*}[t!]
\vspace{-7pt}  %
  \centering
  \renewcommand{\arraystretch}{1.0} 
  \setlength{\tabcolsep}{3pt}

  \caption{Reranking performance under Single- and Mixed-domain settings. All results use Max pooling for audio window size.}
  \label{tab:combined_results}
  \vspace{-5pt}

  \resizebox{0.90\textwidth}{!}{%
  \begin{tabular}{ll ccc ccc c ccc ccc}
    \toprule
    & & \multicolumn{6}{c}{\textbf{Single-domain}} && \multicolumn{6}{c}{\textbf{Mixed-domain}} \\
    \cmidrule(lr){3-8} \cmidrule(lr){10-15}

    & & \multicolumn{3}{c}{\textbf{Spoken SQuAD}} & \multicolumn{3}{c}{\textbf{MS~MARCO}}
    & & \multicolumn{3}{c}{\textbf{Spoken SQuAD}} & \multicolumn{3}{c}{\textbf{MS~MARCO}} \\
    \cmidrule(lr){3-5} \cmidrule(lr){6-8} \cmidrule(lr){10-12} \cmidrule(lr){13-15}

    \textbf{Backbone} & \textbf{Obj.}
       & Hit@1 & MRR & NDCG@5 & Hit@1 & MRR & NDCG@5
      && Hit@1 & MRR & NDCG@5 & Hit@1 & MRR & NDCG@5 \\
    \midrule

    \multicolumn{1}{l}{\textit{Baseline}} 
    & $Z$-score %\\
      & 0.6927 & 0.7498 & 0.7678 & 0.6864 & 0.7694 & 0.7987
      && 0.5273 & 0.6408 & 0.6857 & 0.6621 & 0.7510 & 0.7846 \\
    \midrule

    \sys{Ultravox}
      & \textbf{Pointwise}
         & \textbf{0.7834} & \textbf{0.8008} & \textbf{0.8057} & \textbf{0.7432} & \textbf{0.8060} & \textbf{0.8262}
        && \textbf{0.7630} & \textbf{0.7892} & \textbf{0.7971} & \textbf{0.6977} & \textbf{0.7755} & \textbf{0.8032} \\
      & Pairwise
         & 0.4855 & 0.6288 & 0.6775 & 0.4394 & 0.6195 & 0.6865
        && 0.3602 & 0.5434 & 0.6131 & 0.3336 & 0.5349 & 0.6220 \\
      & Listwise
         & 0.4430 & 0.6017 & 0.6572 & 0.3356 & 0.5487 & 0.6332
        && 0.4430 & 0.6017 & 0.6572 & 0.1185 & 0.3737 & 0.4996 \\
    \midrule

    \sys{Qwen-Audio-Chat}
      & \textbf{Pointwise}
         & \textbf{0.7644} & \textbf{0.7904} & \textbf{0.7980} & 0.7244 & 0.7951 & 0.8180
        && \textbf{0.7334} & \textbf{0.7728} & \textbf{0.7849} & 0.6907 & 0.7720 & 0.8006 \\
      & Pairwise
         & 0.7561 & 0.7861 & 0.7948 & \textbf{0.7279} & \textbf{0.7971} & \textbf{0.8196}
        && 0.7174 & 0.7635 & 0.7780 & \textbf{0.7076} & \textbf{0.7834} & \textbf{0.8092} \\
      & Listwise
         & 0.7361 & 0.7755 & 0.7870 & 0.6967 & 0.7792 & 0.8062
        && 0.7275 & 0.7703 & 0.7831 & 0.5898 & 0.7128 & 0.7564 \\
    \midrule

    \sys{Qwen2-Audio}
      & \textbf{Pointwise}
         & \textbf{0.7753} & \textbf{0.7966} & \textbf{0.8026} & 0.7337 & 0.8004 & 0.8220
        && \textbf{0.7584} & \textbf{0.7872} & \textbf{0.7957} & 0.6997 & 0.7777 & 0.8049 \\
      & Pairwise
         & 0.7705 & 0.7937 & 0.8005 & \textbf{0.7405} & \textbf{0.8047} & \textbf{0.8252}
        && 0.7467 & 0.7803 & 0.7905 & \textbf{0.7164} & \textbf{0.7885} & \textbf{0.8130} \\
      & Listwise
         & 0.7580 & 0.7870 & 0.7955 & 0.6872 & 0.7744 & 0.8027
        && 0.7557 & 0.7858 & 0.7946 & 0.4047 & 0.5976 & 0.6701 \\
    \bottomrule
  \end{tabular}%
  }
\vspace{-10pt}
\end{table*}

\myparagraph{Training Objectives.}
We optimize the reranker using the objective functions detailed in~\autoref{tab:objective}. Our framework is designed for high flexibility, supporting Pointwise (binary cross-entropy)~\cite{liu2009learning, nogueira2020document}, Pairwise (RankNet-style)~\cite{burges2005learning, burges2010ranknet}, and Listwise (ApproxNDCG)~\cite{cao2007learning, xia2008listwise, qin2010general, sharifymoghaddam2025rankllm} loss functions. This modularity allows any of these widely adopted objectives to be seamlessly integrated into our training pipeline.

\myparagraph{Audio Windowing and Score Aggregation.}
To handle potential evidence localization within long audio passages, we segment each candidate into $W$ fixed-duration windows. During training, a single window is randomly sampled for scoring to serve as a stochastic regularizer, thereby enhancing model robustness. In contrast, during inference, the windows are scored independently, with the final passage-level score $s_i$ obtained by aggregating these window-level scores via $\code{mean}(\cdot)$ or $\code{max}(\cdot)$.
\section{Experiment}
\label{sec:setup}

\subsection{Experimental Setup}
This section details the experimental setup and evaluation metrics used in our evaluation.

\myparagraph{Datasets.}
We evaluate \alg using a fixed top-$k$ pipeline ($k=5$) on two distinct datasets: Spoken SQuAD~\cite{li2018spoken} consisting of text queries and TTS-generated audio passages, and MS MARCO~\cite{bajaj2016msmarco}, comprising text-only web passages. The combined retrieval pool contains approximately $2.8$K audio and $9.1$K text passages. These corpora feature non-overlapping passages and minimal query overlap. Given that the datasets differ in both domain and modality, this setup provides a challenging heterogeneous environment to assess the robustness of cross-modal reranking beyond simple modality-based discrimination. 

\myparagraph{Models.}
Our experiments evaluate three audio-native backbone architectures as student rerankers: \sys{Ultravox}~\cite{fixie2024ultravox}, \sys{Qwen-Audio-Chat}~\cite{chu2023qwenaudio}, and \sys{Qwen2-Audio}~\cite{chu2024qwen2}. We compare these three against \textbf{Z-score retrieval} as the primary baseline, which ranks candidates through the modality-wise normalization of retriever scores as described in~\autoref{sec:method}.

\begin{table}[h]
\vspace{-4pt}
\centering
\caption{Evaluation of the foundation ALM. Label quality is measured on $5$K samples, and ranking performance is assessed via Hit@1 on $1$K, respectively.}
\label{tab:gpt_eval}
\vspace{-5pt}
\resizebox{0.85\columnwidth}{!}{
\begin{tabular}{l c c c}
\toprule
\textbf{Category} & \textbf{Metric} & \textbf{Value} & \textbf{Gain ($\Delta$)} \\
\midrule
\multirow{2}{*}{Label Quality} & F1 Score & 0.700 & - \\
                               & MCC & 0.614 & - \\
\midrule
\multirow{3}{*}{Ranking (Hit@1)} & Z-score (Baseline) & 0.578 & - \\
                                 & \textbf{Foundation ALM (gpt-4o)} & \textbf{0.652} & \textbf{+0.074} \\
                                 & Oracle (GT) & 0.825 & +0.247 \\
\bottomrule
\end{tabular}}
\vspace{-10pt}
\end{table}

\myparagraph{Dataset Annotation.}
We use \code{gpt-4o-audio-preview}\footnote{Since \code{gpt-4o-audio-preview} does not support text-only queries, \code{gpt-4o} is used for the candidates that consist solely of text.} for dataset annotation, employing deterministic decoding and structured output to ensure annotation consistency.
To ensure the reliability of the generated labels, we evaluate the \code{gpt-4o-audio-preview} model's performance on a held-out set of $5,000$ candidates. As shown in~\autoref{tab:gpt_eval}, the foundation ALM achieves an F1 score of 0.700 and an MCC (Matthews Correlation Coefficient) of 0.614 against passage-ID ground truth, confirming high-quality label synthesis. Moreover, we assess the ranking signal by evaluating the foundation ALM's direct ranking performance across $1,000$ queries. The annotated labels achieve a Hit@1 of 0.652, outperforming $Z$-score retrieval baseline by $0.074$.

\myparagraph{Training.}
To evaluate the architectural modularity of the proposed algorithm, we train the reranker under three distinct objectives: \emph{Pointwise}, \emph{Pairwise}, and \emph{Listwise} as denoted in~\autoref{tab:objective}. All models are fine-tuned via LoRA with hyperparameter $r=16$, $\alpha=32$, and dropout probability $0.05$ for $3$ epochs using the AdamW optimizer. We set the learning rate to $2\times 10^{-4}$ with $10^{-2}$ weight decay, a $10\%$ linear warmup, and a gradient accumulation factor of $2$. Following the audio windowing strategy, as described in~\autoref{sec:method}, we utilize 30 second segments (\ie $W=4$) within a 120 second total budget.

\myparagraph{Evaluation Scenario.}
To evaluate the model's precision and robustness, we report results in two scenarios based on domain constraints applied after reranking. The \emph{Single-domain} setting is designed to simulate a single-modality environment, ensuring that our approach maintains high performance even with a specific domain by eliminating cross-domain noise. In contrast, the \emph{Mixed-domain} setting evaluates the model against the full heterogeneous pool. This scenario assesses the model's ability to discriminate relevance in complex, multi-modal contexts where candidates from diverse sources are presented simultaneously. 

\myparagraph{Evaluation Metric.}
We assess reranking performance using Hit@1, MRR (Mean Reciprocal Rank), and NDCG@5  (Normalized Discounted Cumulative Gain), which measure the model's ability to correctly identify the ground-truth passage ID within the top-$k$ candidates. These metrics evaluate the accuracy of the ranked list by checking if the target passage is successfully retrieved at the top positions. For downstream QA tasks, we additionally report Exact Match (EM) based on substring matching to evaluate the fidelity of the generated answers against the reference text. Both \emph{Single-domain} and \emph{Mixed-domain} scenarios utilize an identical set of held-out evaluation queries ($\sim$13K from SQuAD and $8.7$K from MS MARCO).

\begin{table*}[t]
\vspace{-7pt}  %
  \centering
  \caption{Downstream QA performance (EM) comparison. \sys{GPT} evaluation is conducted on a representative subset of $1$K samples.}
  \label{tab:generation_em}
  \vspace{-5pt}
  \resizebox{0.86\linewidth}{!}{%
  \begin{tabular}{l l c c @{\hskip 0.4in} l l c c}
    \toprule
    \multicolumn{4}{c}{\textbf{Spoken SQuAD}} & \multicolumn{4}{c}{\textbf{MS MARCO}} \\
    \cmidrule(r{0.4in}){1-4} \cmidrule{5-8}
    \textbf{Generator} & \textbf{Scoring Model} & \textbf{Single} & \textbf{Mixed} & \textbf{Generator} & \textbf{Scoring Model} & \textbf{Single} & \textbf{Mixed} \\
    \midrule
    \sys{Ultravox} & $Z$-score Retrieval & 0.4555 & 0.3565 & \sys{Ultravox} & $Z$-score Retrieval & 0.3637 & 0.3531 \\
             & \sys{Qwen2-Audio} & 0.5046 & 0.4762 &          & \sys{Qwen2-Audio} & 0.3834 & 0.3729 \\
             & \sys{Qwen-Audio-Chat} & 0.4986 & 0.4659 &          & \sys{Qwen-Audio-Chat} & 0.3795 & 0.3666 \\
             & \sys{Ultravox} & \textbf{0.5088} & \textbf{0.4796} &          & \sys{Ultravox} & \textbf{0.3880} & \textbf{0.3719} \\
    \cmidrule(r{0.4in}){1-4} \cmidrule{5-8}
    \sys{Qwen2-Audio} & $Z$-score Retrieval & 0.3335 & 0.2646 & \sys{Qwen2-Audio} & $Z$-score Retrieval & 0.3885 & 0.3774 \\
                & \sys{Qwen2-Audio} & 0.3669 & 0.3468 &             & \sys{Qwen2-Audio} & 0.4090 & \textbf{0.3981} \\
                & \sys{Qwen-Audio-Chat} & 0.3633 & 0.3414 &             & \sys{Qwen-Audio-Chat} & 0.4028 & 0.3898 \\
                & \sys{Ultravox} & \textbf{0.3688} & \textbf{0.3513} &             & \sys{Ultravox} & \textbf{0.4150} & 0.3976 \\
    \cmidrule(r{0.4in}){1-4} \cmidrule{5-8}
    \sys{Qwen-Audio-Chat} & $Z$-score Retrieval & 0.3426 & 0.2794 & \sys{Qwen-Audio-Chat} & $Z$-score Retrieval & 0.3637 & 0.3516 \\
                    & \sys{Qwen2-Audio} & 0.3709 & 0.3499 &                & \sys{Qwen2-Audio} & 0.3798 & \textbf{0.3742} \\
                    & \sys{Qwen-Audio-Chat} & 0.3656 & 0.3455 &                & \sys{Qwen-Audio-Chat} & 0.3761 & 0.3688 \\
                    & \sys{Ultravox} & \textbf{0.3733} & \textbf{0.3555} &                & \sys{Ultravox} & \textbf{0.3873} & 0.3705 \\
    \cmidrule(r{0.4in}){1-4} \cmidrule{5-8}
    \sys{GPT-4o-Audio-Preview} & $Z$-score Retrieval & 0.6273 & 0.4960 & \sys{GPT-4o-Audio-Preview} & $Z$-score Retrieval & 0.3680 & 0.3590 \\
           & \sys{Qwen2-Audio} & 0.6731 & 0.6320 &        & \sys{Qwen2-Audio} & 0.3680 & \textbf{0.3660} \\
           & \sys{Qwen-Audio-Chat} & 0.6731 & 0.6230 &        & \sys{Qwen-Audio-Chat} & 0.3630 & 0.3560 \\
           & \sys{Ultravox} & \textbf{0.6794} & \textbf{0.6350} &        & \sys{Ultravox} & \textbf{0.3820} & 0.3650 \\
    \bottomrule
  \end{tabular}%
  }
\vspace{-10pt}
\end{table*}

\subsection{Results}
\myparagraph{Main results.}
\autoref{tab:combined_results} reports the reranking performance, demonstrating that our approach effectively maintains high precision in a single-domain environment (Single) while exhibiting robust discrimination in heterogeneous contexts (Mixed). In the Single setting, all three backbones successfully identify target information within their native domains, consistently outperforming the $Z$-score baseline when optimized with the appropriate objective. Notably, this performance advantage extends to the Mixed setting, where the models must distinguish relevance across a complex pool of combined speech and text candidates. While the pointwise objective yields the most stable results across backbones, the listwise approach shows higher sensitivity, with \sys{Ultravox} experiencing a significant performance drop ($\Delta$MRR of 0.257) on MS MARCO compared to its pointwise counterpart. 
% These findings confirm that our reranking framework is not only effective for specialized single-modality tasks but also remains highly reliable in multi-modal retrieval scenarios.

\myparagraph{Downstream QA Performance.}
As illustrated in~\autoref{tab:generation_em}, we verify that the improvements in reranking precision directly translate into enhanced end-task performance for downstream Question Answering (QA). By feeding the top-$1$ passage from each pointwise \alg{} into three ALM generators, we observe that our reranking framework consistently boosts substring-match EM over the $Z$-score baseline across all generator-dataset pairs. This enhancement is robustly maintained in both single-domain and mixed-domain scenarios, confirming that the reranker provides high-quality, relevant context that reduces potential hallucinations in generators. Notably, the most substantial gain is recorded in the Spoken SQuAD mixed setting, where the \sys{Ultravox} generator achieves a $0.123$ EM accuracy increase rising from $0.357$ to $0.480$. These results demonstrate the practical utility of our mixed-modality based reranking approach in supporting accurate answer generation within complex, heterogeneous retrieval environments.

\subsection{Analysis}
\begin{table}[ht]
  \vspace{-9pt}
  \centering
  \caption{The ratio of speech evidence in top-5 candidates.}
  \label{tab:modality_collapse}
  \vspace{-5pt}
    \resizebox{0.8\columnwidth}{!}{%
    \setlength{\tabcolsep}{8pt}
    \begin{tabular}{l cc}
      \toprule
      \textbf{Query Set}
        & \textbf{w/o Z-score (\%)}
        & \textbf{w/ Z-score (\%)} \\
      \midrule
      Spoken SQuAD & 4.0 & 45.2 \\
      MS~MARCO     & 0.1 & 18.3 \\
      \bottomrule
    \end{tabular}%
  }
  \vspace{-10pt} 
\end{table}

\myparagraph{Impact of \textbf{Z-score Aggregation.}}
As described in~\autoref{tab:modality_collapse}, to isolate retrieval-stage fusion effects, we compare unnormalized fusion, modality-wise $Z$-score, and Reciprocal Rank Fusion (RRF). On Spoken SQuAD, both unnormalized fusion and RRF suffer from modality collapse, yielding near-zero Hit@1 ($\approx0.05$) as raw text scores overwhelm speech candidates. $Z$-score normalization effectively resolves this bias, significantly recovering Hit@1 to $0.527$. While unnormalized fusion slightly outperforms $Z$-score on MS MARCO ($0.687$ vs. $0.662$) due to the inherent dominance of text candidates, we adopt $Z$-score as the default. Unlike other methods, $Z$-score consistently ensures speech visibility across both domains, preventing the exclusion of speech candidates in mixed modality retrieval.

\begin{table}[ht]
  \vspace{-4pt}
  \centering
  \caption{MRR under inference-time Mean/Max pooling.}
  \label{tab:agg}
  \vspace{-5pt}
  
  \resizebox{0.8\columnwidth}{!}{%
    \setlength{\tabcolsep}{6pt}
    \begin{tabular}{l cc cc}
      \toprule
      \textbf{Backbone}
        & \multicolumn{2}{c}{\textbf{MS~MARCO}}
        & \multicolumn{2}{c}{\textbf{Spoken SQuAD}} \\
      \cmidrule(lr){2-3} \cmidrule(lr){4-5}
        & \textbf{Mean} & \textbf{Max}
        & \textbf{Mean} & \textbf{Max} \\
      \midrule
      \sys{Ultravox}
        & \textbf{0.7918} & 0.7755 & 0.7460 & \textbf{0.7892} \\
      \sys{Qwen-Audio-Chat}
        & \textbf{0.7821} & 0.7720 & 0.7286 & \textbf{0.7728} \\
      \sys{Qwen2-Audio}
        & \textbf{0.7885} & 0.7777 & 0.7468 & \textbf{0.7872} \\
      \bottomrule
    \end{tabular}%
  }
  \vspace{-8pt}
\end{table}

\myparagraph{Analysis of Pooling Strategy.}
\autoref{tab:agg} compares \emph{Mean} and \emph{Max} pooling strategies for aggregating speech window scores in the pointwise Mixed setting. The choice of pooling primarily affects audio-heavy contexts; on MS MARCO, where the audio candidate share is relatively low, the performance gap is marginal, with Mean pooling holding a slight edge. However, on Spoken SQuAD where speech candidates constitute a larger portion of the top-$5$ pool,  Max pooling consistently outperforms Mean across all backbone cases.

\begin{table}[h]
  \vspace{-3pt}
  \centering
  \caption{Window-length analysis on \sys{Qwen2-Audio} (Mixed).}
  \label{tab:window}
  \vspace{-5pt}
  \resizebox{0.85\columnwidth}{!}{%
    \setlength{\tabcolsep}{3.5pt}
    \begin{tabular}{l ccc ccc}
      \toprule
      & \multicolumn{3}{c}{\textbf{Spoken SQuAD}}
        & \multicolumn{3}{c}{\textbf{MS~MARCO}} \\
      \cmidrule(lr){2-4} \cmidrule(lr){5-7}
      \textbf{Window}
        & Hit@1 & MRR & NDCG@5
        & Hit@1 & MRR & NDCG@5 \\
      \midrule
      120s${\times}$1
        & 0.7509 & 0.7822 & 0.7919
        & 0.7092 & \textbf{0.7834} & 0.8091 \\
      60s${\times}$2
        & 0.7515 & 0.7827 & 0.7922
        & \textbf{0.7138} & 0.7862 & \textbf{0.8112} \\
      30s${\times}$4
        & \textbf{0.7584} & \textbf{0.7872} & \textbf{0.7957}
        & 0.6997 & 0.7777 & 0.8049 \\
      \bottomrule
    \end{tabular}%
  }
  \vspace{-10pt} 
\end{table}
\myparagraph{Window-length Analysis.}
\autoref{tab:window} compares three window configurations on \sys{Qwen2-Audio} under a fixed 120 seconds audio budget and Max pooling. Given the inherent maximum sequence length constraints of the model's audio encoder, segmenting long audio into multiple windows is necessary to capture full temporal information without truncation. On Spoken SQuAD, finer segmentation (30s$\times$4) achieves the highest Hit@1 score (0.758), outperforming the single-window (120s$\times$1). While the performance gap remains marginal on MS MARCO due to the lower audio candidate density, these results indicate that finer-grained windows can more effectively pinpoint salient information while operating within the model's architectural limits. 
\section{Conclusion} 
\label{sec:conclusion} 
We investigated reranking in heterogeneous pools of speech and text, demonstrating that \alg effectively maintains high precision in a single-domain environment while ensuring robust discrimination in mixed-modality scenarios. By addressing structural challenges through modality-wise $Z$-score normalization and an optimized audio windowing strategy to overcome sequence length constraints, we achieve consistent performance gains. Our findings highlight that the pointwise objective is the primary driver of stability, whereas listwise approaches can lead to sharp degradation in heterogeneous settings. Ultimately, these reranking improvements translate into enhanced downstream QA performance, providing reliable context for accurate answer generation. While this study utilizes TTS-based data, \ie Spoken SQuAD, future work will focus on validating these insights with natural, spontaneous speech.

\section{Acknowledgments}
This work was supported by the Institute of Information \& Communications Technology Planning \& Evaluation (IITP) grant funded by the Korea government (MSIT) (RS-2025-25464461, AI's Vision of Harmony: A Fair and Transparent Multimodal Agentic Platform for Conflict Mediation)

\section{Generative AI Use Disclosure}
Large Language Model assistance was used for language editing and polishing of portions of this manuscript. Beyond the specific experimental procedures explicitly described in the methodology, such as the use of generative models for synthetic dataset construction, no generative AI tool was used to produce experimental results, figures, tables or the scientific content of this work.

\bibliographystyle{IEEEtran}
\bibliography{ref}

@inproceedings{elizalde2023clap,
  title={Clap learning audio concepts from natural language supervision},
  author={Elizalde, Benjamin and Deshmukh, Soham and Al Ismail, Mahmoud and Wang, Huaming},
  booktitle={ICASSP 2023-2023 IEEE International Conference on Acoustics, Speech and Signal Processing (ICASSP)},
  pages={1--5},
  year={2023},
  organization={IEEE}
}

@article{feng2025enhancing,
  title={Enhancing Speech-to-Speech Dialogue Modeling with End-to-End Retrieval-Augmented Generation},
  author={Feng, Pengchao and Ma, Ziyang and Chen, Wenxi and Li, Yao and Wang, Sheng and Yu, Kai and Chen, Xie},
  journal={arXiv preprint arXiv:2505.00028},
  year={2025}
}

@inproceedings{saijo2025leveraging,
  title={Leveraging audio-only data for text-queried target sound extraction},
  author={Saijo, Kohei and Ebbers, Janek and Germain, Fran{\c{c}}ois G and Khurana, Sameer and Wichern, Gordon and Le Roux, Jonathan},
  booktitle={ICASSP 2025-2025 IEEE International Conference on Acoustics, Speech and Signal Processing (ICASSP)},
  pages={1--5},
  year={2025},
  organization={IEEE}
}

@inproceedings{sun2023chatgpt,
  title={Is ChatGPT good at search? investigating large language models as re-ranking agents},
  author={Sun, Weiwei and Yan, Lingyong and Ma, Xinyu and Wang, Shuaiqiang and Ren, Pengjie and Chen, Zhumin and Yin, Dawei and Ren, Zhaochun},
  booktitle={Proceedings of the 2023 conference on empirical methods in natural language processing},
  pages={14918--14937},
  year={2023}
}

@inproceedings{qin2024large,
  title={Large language models are effective text rankers with pairwise ranking prompting},
  author={Qin, Zhen and Jagerman, Rolf and Hui, Kai and Zhuang, Honglei and Wu, Junru and Yan, Le and Shen, Jiaming and Liu, Tianqi and Liu, Jialu and Metzler, Donald and others},
  booktitle={Findings of the Association for Computational Linguistics: NAACL 2024},
  pages={1504--1518},
  year={2024}
}

@misc{bajaj2016msmarco,
      title={MS MARCO: A Human Generated MAchine Reading COmprehension Dataset}, 
      author={Payal Bajaj and Daniel Campos and Nick Craswell and Li Deng and Jianfeng Gao and Xiaodong Liu and Rangan Majumder and Andrew McNamara and Bhaskar Mitra and Tri Nguyen and Mir Rosenberg and Xia Song and Alina Stoica and Saurabh Tiwary and Tong Wang},
      year={2018},
      eprint={1611.09268},
      archivePrefix={arXiv},
      primaryClass={cs.CL},
      url={https://arxiv.org/abs/1611.09268}, 
}

@inproceedings{burges2005learning,
  title={Learning to rank using gradient descent},
  author = {Burges, Chris and Shaked, Tal and Renshaw, Erin and Lazier, Ari and Deeds, Matt and Hamilton, Nicole and Hullender, Greg},
  year = {2005},
  isbn = {1595931805},
  publisher = {Association for Computing Machinery},
  address = {New York, NY, USA},
  url = {https://doi.org/10.1145/1102351.1102363},
  doi = {10.1145/1102351.1102363},
  booktitle = {Proceedings of the 22nd International Conference on Machine Learning},
  pages = {89–96},
  numpages = {8},
  location = {Bonn, Germany},
  series = {ICML '05}
}

@misc{chu2023qwenaudio,
      title={Qwen-Audio: Advancing Universal Audio Understanding via Unified Large-Scale Audio-Language Models}, 
      author={Yunfei Chu and Jin Xu and Xiaohuan Zhou and Qian Yang and Shiliang Zhang and Zhijie Yan and Chang Zhou and Jingren Zhou},
      year={2023},
      eprint={2311.07919},
      archivePrefix={arXiv},
      primaryClass={eess.AS},
      url={https://arxiv.org/abs/2311.07919}, 
}

@misc{fixie2024ultravox,
  title={Ultravox: A Fast Multimodal LLM for Real-Time Voice},
  author={{Fixie AI}},
  year={2024},
  howpublished={\url{https://github.com/fixie-ai/ultravox}},
  note={Accessed: 2025}
}

@misc{min2024speech,
  title={Speech Retrieval-Augmented Generation without Automatic Speech Recognition},
  author={Do June Min and Karel Mundnich and Andy Lapastora and Erfan Soltanmohammadi and Srikanth Ronanki and Kyu Han},
  year={2024},
  eprint={2412.16500},
  archivePrefix={arXiv},
  primaryClass={eess.AS},
  url={https://arxiv.org/abs/2412.16500}
}

@inproceedings{rackauckas2025voxrag,
    title = "{V}ox{RAG}: A Step Toward Transcription-Free {RAG} Systems in Spoken Question Answering",
    author = "Rackauckas, Zackary  and
      Hirschberg, Julia",
    editor = "Kriz, Reno  and
      Murray, Kenton",
    booktitle = "Proceedings of the 1st Workshop on Multimodal Augmented Generation via Multimodal Retrieval (MAGMaR 2025)",
    month = aug,
    year = "2025",
    address = "Vienna, Austria",
    publisher = "Association for Computational Linguistics",
    url = "https://aclanthology.org/2025.magmar-1.3/",
    doi = "10.18653/v1/2025.magmar-1.3",
    pages = "40--46",
    ISBN = "979-8-89176-280-0"
}

@inproceedings{chen2025wavrag,
  title={WavRAG: Audio-Integrated Retrieval Augmented Generation for Spoken Dialogue Models},
  author={Yifu Chen and Shengpeng Ji and Haoxiao Wang and Ziqing Wang and Siyu Chen and Jinzheng He and Jin Xu and Zhou Zhao},
  booktitle={Proceedings of the 63rd Annual Meeting of the Association for Computational Linguistics (Volume 1: Long Papers)},
  pages={12505--12523},
  year={2025},
  address={Vienna, Austria},
  publisher={Association for Computational Linguistics},
  doi={10.18653/v1/2025.acl-long.613},
  url={https://arxiv.org/abs/2502.14727}
}

@inproceedings{hu2022lora,
  title={{LoRA}: Low-rank adaptation of large language models},
  author={Hu, Edward J and Shen, Yelong and Wallis, Phillip and Allen-Zhu, Zeyuan and Li, Yuanzhi and Wang, Shean and Wang, Lu and Chen, Weizhu},
  booktitle={International Conference on Learning Representations (ICLR)},
  year={2022}
}

@article{li2018spoken,
  title={Spoken squad: A study of mitigating the impact of speech recognition errors on listening comprehension},
  author={Li, Chia-Hsuan and Wu, Szu-Lin and Liu, Chi-Liang and Lee, Hung-yi},
  journal={arXiv preprint arXiv:1804.00320},
  year={2018}
}

@inproceedings{wang2023e5mistral,
  title={Improving text embeddings with large language models},
  author={Wang, Liang and Yang, Nan and Huang, Xiaolong and Yang, Linjun and Majumder, Rangan and Wei, Furu},
  booktitle={Proceedings of the 62nd Annual Meeting of the Association for Computational Linguistics (Volume 1: Long Papers)},
  pages={11897--11916},
  year={2024}
}

@inproceedings{xia2008listwise,
  title={Listwise approach to learning to rank: theory and algorithm},
  author={Xia, Fen and Liu, Tie-Yan and Wang, Jue and Zhang, Wensheng and Li, Hang},
  booktitle={Proceedings of the 25th international conference on Machine learning},
  pages={1192--1199},
  year={2008}
}

@article{qin2010general,
  title={A general approximation framework for direct optimization of information retrieval measures},
  author={Qin, Tao and Liu, Tie-Yan and Li, Hang},
  journal={Information retrieval},
  volume={13},
  number={4},
  pages={375--397},
  year={2010},
  publisher={Springer}
}

@article{achiam2023gpt,
  title={Gpt-4 technical report},
  author={Achiam, Josh and Adler, Steven and Agarwal, Sandhini and Ahmad, Lama and Akkaya, Ilge and Aleman, Florencia Leoni and Almeida, Diogo and Altenschmidt, Janko and Altman, Sam and Anadkat, Shyamal and others},
  journal={arXiv preprint arXiv:2303.08774},
  year={2023}
}

@article{chu2024qwen2,
  title={Qwen2-audio technical report},
  author={Chu, Yunfei and Xu, Jin and Yang, Qian and Wei, Haojie and Wei, Xipin and Guo, Zhifang and Leng, Yichong and Lv, Yuanjun and He, Jinzheng and Lin, Junyang and others},
  journal={arXiv preprint arXiv:2407.10759},
  year={2024}
}

@article{team2023gemini,
  title={Gemini: a family of highly capable multimodal models},
  author={Team, Gemini and Anil, Rohan and Borgeaud, Sebastian and Alayrac, Jean-Baptiste and Yu, Jiahui and Soricut, Radu and Schalkwyk, Johan and Dai, Andrew M and Hauth, Anja and Millican, Katie and others},
  journal={arXiv preprint arXiv:2312.11805},
  year={2023}
}

@article{lewis2020retrieval,
  title={Retrieval-augmented generation for knowledge-intensive nlp tasks},
  author={Lewis, Patrick and Perez, Ethan and Piktus, Aleksandra and Petroni, Fabio and Karpukhin, Vladimir and Goyal, Naman and K{\"u}ttler, Heinrich and Lewis, Mike and Yih, Wen-tau and Rockt{\"a}schel, Tim and others},
  journal={Advances in neural information processing systems},
  volume={33},
  pages={9459--9474},
  year={2020}
}

@article{liu2009learning,
  title={Learning to rank for information retrieval},
  author={Liu, Tie-Yan},
  journal={Foundations and Trends{\textregistered} in Information Retrieval},
  volume={3},
  number={3},
  pages={225--331},
  year={2009},
  publisher={Emerald Publishing Limited}
}

@inproceedings{nogueira2020document,
  title={Document ranking with a pretrained sequence-to-sequence model},
  author={Nogueira, Rodrigo and Jiang, Zhiying and Pradeep, Ronak and Lin, Jimmy},
  booktitle={Findings of the association for computational linguistics: EMNLP 2020},
  pages={708--718},
  year={2020}
}

@article{burges2010ranknet,
  title={From ranknet to lambdarank to lambdamart: An overview},
  author={Burges, Christopher JC},
  journal={Learning},
  volume={11},
  number={23-581},
  pages={81},
  year={2010}
}

@inproceedings{cao2007learning,
  title={Learning to rank: from pairwise approach to listwise approach},
  author={Cao, Zhe and Qin, Tao and Liu, Tie-Yan and Tsai, Ming-Feng and Li, Hang},
  booktitle={Proceedings of the 24th international conference on Machine learning},
  pages={129--136},
  year={2007}
}

@article{huang2025survey,
  title={A survey on hallucination in large language models: Principles, taxonomy, challenges, and open questions},
  author={Huang, Lei and Yu, Weijiang and Ma, Weitao and Zhong, Weihong and Feng, Zhangyin and Wang, Haotian and Chen, Qianglong and Peng, Weihua and Feng, Xiaocheng and Qin, Bing and others},
  journal={ACM Transactions on Information Systems},
  volume={43},
  number={2},
  pages={1--55},
  year={2025},
  publisher={ACM New York, NY}
}

@article{liang2022mind,
  title={Mind the gap: Understanding the modality gap in multi-modal contrastive representation learning},
  author={Liang, Victor Weixin and Zhang, Yuhui and Kwon, Yongchan and Yeung, Serena and Zou, James Y},
  journal={Advances in Neural Information Processing Systems},
  volume={35},
  pages={17612--17625},
  year={2022}
}

@inproceedings{sharifymoghaddam2025rankllm,
  title={Rankllm: A python package for reranking with llms},
  author={Sharifymoghaddam, Sahel and Pradeep, Ronak and Slavescu, Andre and Nguyen, Ryan and Xu, Andrew and Chen, Zijian and Zhang, Yilin and Chen, Yidi and Xian, Jasper and Lin, Jimmy},
  booktitle={Proceedings of the 48th International ACM SIGIR Conference on Research and Development in Information Retrieval},
  pages={3681--3690},
  year={2025}
}

@inproceedings{schlatt2025set,
  title={Set-encoder: Permutation-invariant inter-passage attention for listwise passage re-ranking with cross-encoders},
  author={Schlatt, Ferdinand and Fr{\"o}be, Maik and Scells, Harrisen and Zhuang, Shengyao and Koopman, Bevan and Zuccon, Guido and Stein, Benno and Potthast, Martin and Hagen, Matthias},
  booktitle={European Conference on Information Retrieval},
  pages={1--19},
  year={2025},
  organization={Springer}
}

@inproceedings{yoon2024listt5,
  title={Listt5: Listwise reranking with fusion-in-decoder improves zero-shot retrieval},
  author={Yoon, Soyoung and Choi, Eunbi and Kim, Jiyeon and Yun, Hyeongu and Kim, Yireun and Hwang, Seung-won},
  booktitle={Proceedings of the 62nd Annual Meeting of the Association for Computational Linguistics (Volume 1: Long Papers)},
  pages={2287--2308},
  year={2024}
}

\end{document}